%% file: main.tex
\documentclass{article}
\usepackage{arxiv}
\usepackage[utf8]{inputenc}
\usepackage[T1]{fontenc}
\usepackage{amsmath,amssymb,mathtools}
\usepackage{newtxtext,newtxmath}
\usepackage[hyphens]{url}
\usepackage{graphicx,booktabs,array}
\usepackage{microtype}
\usepackage[numbers,sort&compress]{natbib}
\usepackage[font=small,labelfont=bf]{caption}
\usepackage[section]{placeins}
\usepackage{flafter}
\usepackage{algorithm,algorithmic}
\usepackage{hyperref}
\hypersetup{
  hidelinks,
  pdftitle={Learning Collective Dynamics with Differentiable Gaussian Representations},
  pdfkeywords={collective dynamics, Gaussian representations, aggregate observations, response prediction, differentiable learning}
}
\graphicspath{{figures/}}
\DeclareMathOperator{\softmax}{softmax}
\DeclareMathOperator{\softplus}{softplus}
\DeclareMathOperator{\clip}{clip}
\DeclareMathOperator{\NB}{NB}
\newcommand{\E}{\mathbb E}

\title{Learning Collective Dynamics\\with Differentiable Gaussian Representations}
\renewcommand{\shorttitle}{Learning Collective Dynamics}
\input{authors}
\date{}

\begin{document}
\maketitle
\input{abstract}
\input{sections/01-introduction}
\input{sections/02-related-work}
\input{sections/03-method}
\input{sections/04-experiments}

\input{sections/05-conclusion}

\section*{Statement on the Use of AI Tools}
AI tools assisted with manuscript organization, language editing, and preparation of the LaTeX source and diagrams. The reported experiments and numerical results derive from the research artifacts underlying this work. The authors are responsible for the content of the paper.

\clearpage
\bibliographystyle{unsrtnat}
\bibliography{references}

\clearpage
\appendix
\numberwithin{equation}{section}
\input{appendix}

\end{document}

%% file: authors.tex
\author{%
  \normalfont
  \textbf{Jianxiang Ma\textsuperscript{3,1}, Mingfu Zhang\textsuperscript{1,2,*}, Xiaocui Yang\textsuperscript{3}}\\
  \textbf{Yichen Gao\textsuperscript{3}, Junzhao Huang\textsuperscript{3}, Yuesong Hou\textsuperscript{3}}\\
  \textsuperscript{1}OranAI, Shenzhen 518057, China\\
  \textsuperscript{2}OranAI Ltd., City of Industry, CA 91748, USA\\
  \textsuperscript{3}School of Computer Science and Engineering,\\
  Northeastern University, Shenyang 110819, China\\
  \texttt{jianxiangma020518@gmail.com}\\
  \textsuperscript{*}Corresponding author: \texttt{cto@oran.cn}%
}
\hypersetup{pdfauthor={Jianxiang Ma, Mingfu Zhang, Xiaocui Yang, Yichen Gao, Junzhao Huang, Yuesong Hou}}

%% file: abstract.tex
\begin{abstract}
Collective responses depend on individual differences, contact opportunities, and accumulated experience. Learning their dynamics from aggregate counts requires connecting a population's response distribution to both current observations and future behavior. We introduce \textbf{Differentiable Gaussian Dynamics (DGD)}, which learns this connection through three components: a Gaussian mixture representing heterogeneous response propensities, differentiable aggregation of contact intensity and behavioral probabilities, and feedback recurrence that updates subsequent responses. Reparameterized integration and temporal recurrence let aggregate prediction errors jointly train the distribution, observation functions, and feedback parameters. On four windows from KuaiRand-Pure and Online Retail II, DGD achieves lower joint behavioral negative log-likelihood than a DeepAR adaptation with a joint-behavior head. In Retail 2010, its one-day behavioral-count MAE is 4.71 versus 6.88 for this adaptation. Learning the distribution reduces behavioral negative log-likelihood by 10.82\% relative to a fixed Gaussian in KuaiRand's standard-recommendation window; removing feedback dynamics raises joint KL from 0.0340 to 0.2577 in a controlled experiment. These results establish the value of learning population representations and their feedback process from aggregate observations. Code is available at \url{https://github.com/OranAi-Ltd/oransim}.
\end{abstract}

%% file: sections/01-introduction.tex
\section{Introduction}
\label{sec:introduction}

Social-media interactions and market transactions are collective outcomes of individual responses \citep{bass1969new,salganik2006experimental}. Their evolution depends on who participates, how frequently participants encounter an opportunity, and what they have experienced. Granovetter's account of collective behavior shows how distributions of individual thresholds determine group outcomes \citep{granovetter1978threshold}. For prediction, this raises a learning problem: how can aggregate event counts reveal the heterogeneous responses and feedback process that generate future collective behavior? Three coupled challenges determine the model this problem requires. \textbf{Response heterogeneity:} two audiences with the same mean propensity can have different response rates because behavioral probabilities are nonlinear; both differences between response types and variation within each type matter \citep{allenby1998marketing,train2009discrete}. \textbf{Contact-dependent observations:} frequently reached participants contribute more events, so the response distribution in a log depends jointly on population composition and contact intensity \citep{liang2016modeling,schnabel2016recommendations,gao2022kuairand}. \textbf{Experience-dependent dynamics:} current contacts and responses accumulate as memory and fatigue \citep{kronrod2019ad,kapoor2015just}, changing the distribution that will generate later observations \citep{chaney2018algorithmic,jiang2019degenerate}. Aggregate counts combine these effects, so learning from them requires a path from the population state to current observations and from those observations to the next state.

Existing methods provide foundations for this task: mixture choice models integrate over preferences to predict aggregate choices \citep{train2009discrete,berry1995automobile}; aggregate dynamic models learn latent states and their evolution from population observations \citep{sheldon2011collective,singh2022aggregatehmm,singh2023aggregategaussian}; world models predict subsequent observations through state transitions \citep{ha2018worldmodels,hafner2025mastering}; and differentiable simulation trains behavioral or transition parameters from observation errors \citep{mladenov2021recsimng,chopra2023gradabm}. Explicit Gaussian representations suggest a way to construct this path. In 3D Gaussian Splatting, differentiable image synthesis connects Gaussian scene parameters to observed pixels, allowing image errors to train the representation \citep{kerbl2023gaussiansplatting}. We apply this representation--observation learning principle to collective responses: Gaussian components describe latent response propensities, and integration synthesizes observable population counts. As multiple image views constrain one scene, counts from different dates and observation units constrain a base response distribution shared across dates. Feedback then carries the consequences of current responses into the next prediction.

We introduce \textbf{Differentiable Gaussian Dynamics (DGD)}, with one component for each challenge. For response heterogeneity, a \textbf{Gaussian response representation} models type proportions, typical propensities, and variation within types through mixture weights, means, and covariances. For contact-dependent observations, \textbf{differentiable contact--behavior aggregation} evaluates contact intensity and behavioral probabilities at the same latent propensity, then integrates their product to predict arrivals (logged exposures or retail invoices) and behavioral totals. For experience-dependent dynamics, \textbf{feedback recurrence} uses observed or predicted outcomes to update memory, fatigue, response shifts, and the arrival environment. The components form one trainable model: integration propagates current prediction errors to the population representation, while recurrence lets later errors train the feedback process. During training, an expected-feedback branch also advances the state with predicted counts, as forecasting does.

We evaluate DGD on standard-recommendation and random-exposure windows from KuaiRand-Pure and two annual windows from Online Retail II. DGD has lower joint behavioral negative log-likelihood (NLL) than DeepAR-joint, a DeepAR adaptation with a joint-behavior head \citep{salinas2020deepar}, in all four windows, including 1.1232 versus 1.3466 under standard recommendation. Its one-day behavioral-count MAE is lower than DeepAR-joint and three historical-distribution baselines in both retail windows. Learning the distribution lowers standard-recommendation NLL from 1.2594 to 1.1232 relative to a fixed Gaussian, and feedback recurrence reduces random-exposure one-day behavioral-count MAE from 34.08 to 16.47. Controlled experiments recover known cross-day response distributions and predict choice distributions under new product attributes and prices.

Our contributions are:
\begin{itemize}
  \item \textbf{A collective-dynamics model with an explicit population representation.} DGD connects a Gaussian response distribution, contact--behavior aggregation, and feedback recurrence to predict current and future collective responses.
  \item \textbf{Joint learning of representation and dynamics from aggregate counts.} Reparameterized integration and observed- and expected-feedback training propagate prediction losses to distribution, observation, and transition parameters within one objective.
  \item \textbf{Predictive evaluation and analysis of aggregate learning.} Four real-data windows and controlled experiments evaluate behavioral prediction, distribution learning, and feedback; theoretical and numerical analyses establish count-likelihood sufficiency, Gaussian parameter equivalence, and propagation of quadrature error through time.
\end{itemize}

%% file: sections/02-related-work.tex
\section{Related Work}
\label{sec:related}

\subsection{Learning Collective Responses from Population Observations}
Threshold models connect individual differences to collective outcomes \citep{granovetter1978threshold}, while mixture choice models integrate conditional responses over continuous preference distributions or discrete latent classes \citep{train2009discrete,mcfadden2000mixed}. \citet{berry1995automobile} estimate random-coefficient preference distributions from market shares, and \citet{lu2026sparsedemand} combine random-coefficient integration, market totals, and count likelihoods for demand estimation. These formulations motivate learning response distributions from population observations. Collective graphical models learn individual-level models from aggregate counts \citep{sheldon2011collective}, and learning from aggregate observations trains instance-level predictors from group-level supervision \citep{zhang2020learning}. Aggregate hidden Markov models learn latent dynamics through aggregate forward--backward inference and expectation--maximization \citep{singh2022aggregatehmm,singh2023aggregategaussian}, and population-dynamics models learn how distributions evolve from unpaired snapshots \citep{hashimoto2016learning,tong2020trajectorynet,bunne2022proximal}. Neural Hawkes models learn history-dependent event intensities \citep{mei2017neuralhawkes}, and world models learn representations and transitions for predicting future observations \citep{ha2018worldmodels,hafner2025mastering}; Appendix~\ref{app:related} reviews further event-intensity, count-forecasting, and simulation models. RecSim NG supports latent-variable learning in recommender ecosystems \citep{mladenov2021recsimng}; GradABM calibrates behavioral mechanisms through gradients from aggregate outputs \citep{chopra2023gradabm}, and differentiable agent-based simulation further develops gradient computation and aggregate supervision \citep{andelfinger2021differentiable,querabofarull2025automatic}. OranSim uses platform outcome predictions and population rollouts to select marketing scenarios \citep{ma2026oransimoutcome}. DGD uses a latent propensity shared by contact intensity and behavior and jointly learns its distribution, observation functions, and feedback transitions from daily count and behavioral-pattern losses.

\subsection{Explicit Gaussian Representations}
3D Gaussian Splatting optimizes explicit scene components through differentiable rendering \citep{kerbl2023gaussiansplatting}. Dynamic Gaussian representations move or deform components over time to model changing scenes \citep{luiten2024dynamic,wu2024gaussian4d}. Gaussian world models predict scene states following robot actions \citep{lu2025gaussianworldmodel}, and ContactGaussian-WM learns physical parameters using Gaussian geometry, contact dynamics, and rendering \citep{wang2026contactgaussian}. DGD's Gaussian components occupy response-propensity space. Differentiable integration maps this representation to population counts, and response feedback updates its subsequent state.

%% file: sections/03-method.tex
\section{Differentiable Gaussian Dynamics}
\label{sec:method}

\begin{figure}[!t]
  \centering
  \includegraphics[width=\linewidth]{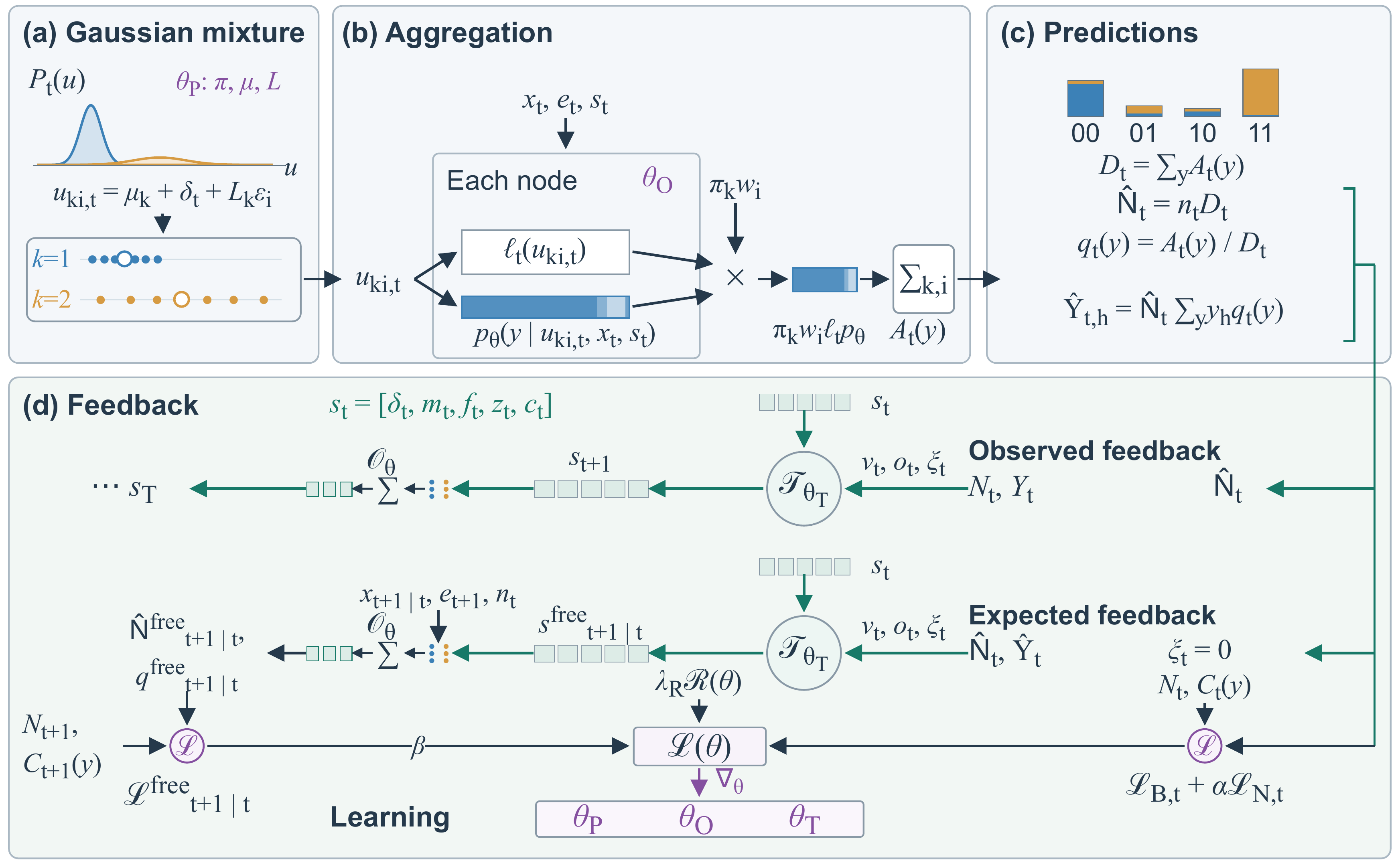}
  \caption{\textbf{Differentiable Gaussian Dynamics (DGD).} (a) The state translates a learned Gaussian mixture. (b) Arrival intensity and joint behavioral probabilities are multiplied at each node, then weighted and summed into pattern intensities $A_t(y)$. (c) Total intensity $D_t=\sum_y A_t(y)$ yields $\widehat N_t=n_tD_t$, $q_t(y)=A_t(y)/D_t$, and $\widehat Y_{t,h}=\widehat N_t\sum_y y_hq_t(y)$. (d) Observed and expected feedback advance the same initial state using recorded and predicted counts, respectively. Current and next-day expected-feedback losses train the population ($\theta_P$), observation ($\theta_O$), and transition ($\theta_T$) parameters. Curves, nodes, and bars in (a)--(c) use a Retail-2010 fit, displaying two behaviors after marginalizing the third; unit and cohort indices are omitted.}
  \label{fig:method}
\end{figure}

DGD has three components: a \emph{Gaussian response representation} for heterogeneous propensities, \emph{contact--behavior aggregation} for generating observations, and \emph{feedback recurrence} for advancing the state (Figure~\ref{fig:method}). Sections~\ref{sec:population}--\ref{sec:recurrence} develop them in this order. Section~\ref{sec:learning} explains how aggregate losses train all three jointly.

\paragraph{Prediction task.}
An observation unit $j\in\{1,\ldots,J\}$ contains disjoint cohorts $g\in\{1,\ldots,G\}$. Units are creator streams in the social-media task and the whole store in the retail task. On date $t$, known features $x_{jgt}$, reference cohort size $n_{jgt}$, and nonnegative activity effort $e_{jgt}$ accompany the observations. An arrival is a logged exposure in the social-media task and a retained merchandise invoice in the retail task; each arrival carries $H$ binary behaviors $y\in\{0,1\}^H$. Daily pattern frequencies $C_{jgt}(y)$ give arrival counts $N_{jgt}=\sum_y C_{jgt}(y)$ and behavior counts $Y_{jgth}=\sum_y y_hC_{jgt}(y)$. From preceding history and known future inputs, DGD predicts expected arrivals $\widehat N_{jgt}$, joint probabilities $q_{jgt}(y)$ conditional on arrival, and expected behavior counts $\widehat Y_{jgth}$. The beginning-of-day state $s_t$ is used to predict day $t$ before that day's outcomes update $s_{t+1}$.

\subsection{Gaussian Response Representation}
\label{sec:population}

We represent differences between and within response types with a $K$-component Gaussian mixture over latent propensity $U\in\mathbb R^d$:
\begin{equation}
P_{g,t}(u)=\sum_{k=1}^{K}\pi_{gk}\mathcal N(u;\mu_{gk}+\delta_{g,t},\Sigma_{gk}),
\quad \pi_g=\softmax(\omega_g),\quad \Sigma_{gk}=L_{gk}L_{gk}^{\top},
\label{eq:population}
\end{equation}
where $\omega_g$ are mixture logits, $\pi_{gk}$ is type $k$'s population share, $\mu_{gk}$ its base mean, and $\Sigma_{gk}$ its covariance. These parameters, denoted by $\theta_P$, are learned and shared across dates. Feedback changes the cohort shift $\delta_{g,t}$, translating its components together. The lower-triangular factor $L_{gk}$ has positive diagonals parameterized by softplus plus a positive floor \citep{pinheiro1996unconstrained}.

To make predictions differentiable with respect to the distribution, we transform fixed standard-normal integration nodes $\epsilon_i$ into component nodes by reparameterization \citep{kingma2014auto,rezende2014stochastic}:
\begin{equation}
u_{gkit}=\mu_{gk}+\delta_{g,t}+L_{gk}\epsilon_i.
\label{eq:nodes}
\end{equation}
Node locations move with the learned means and covariances, so errors in predictions evaluated at these nodes produce pathwise gradients for the representation \citep{mohamed2020monte}. Each node has a positive weight $w_i$ with $\sum_iw_i=1$. We use Gauss--Hermite quadrature \citep{golub1969calculation} in one or two dimensions and fixed reparameterized Sobol nodes \citep{sobol1967distribution,buchholz2018quasi} in higher dimensions.

\subsection{Differentiable Contact--Behavior Aggregation}
\label{sec:aggregation}

A type's contribution to recorded behavior depends on both its contact opportunities and its response after contact \citep{liang2016modeling,ma2018entire}. DGD evaluates these quantities at the same latent propensity $u$. Per-person arrival intensity is
\begin{equation}
\ell_{jgt}(u)=e_{jgt}\exp\{\clip(a_{jg}+\beta_I^{\top}x_{jgt}+z_t+c_{jt}+\gamma^{\top}u,-16,10)\},
\label{eq:intensity}
\end{equation}
where $a_{jg}$ combines baseline log-intensity and a learned intercept, $\beta_I$ gives feature effects, $z_t$ is the common arrival environment, and $c_{jt}$ is the unit deviation. The learned arrival loading $\gamma$ is shared across contexts. The operator $\clip$ restricts the log-intensity before the effort multiplier to $[-16,10]$. Features are standardized using training-period statistics.

Conditional behavior follows an autoregressive logistic model \citep{neal1992connectionist,larochelle2011neural}:
\begin{equation}
\begin{aligned}
p_\theta(y\mid u,x,s_t)&=\prod_{h=1}^{H}\sigma(\eta_h)^{y_h}[1-\sigma(\eta_h)]^{1-y_h},\\
\eta_h&=b_{jgh}(x,s_t)+\lambda_h^{\top}u+\sum_{h'<h}\Psi_{hh'}y_{h'},
\end{aligned}
\label{eq:behavior}
\end{equation}
where $\sigma$ is the logistic function and $\eta_h$ is behavior $h$'s logit. The intercept function $b_{jgh}$ contains unit and cohort intercepts, feature effects, and linear effects of memory, fatigue, and the common environment. The loading vector $\lambda_h$ maps latent propensity to behavior $h$. We fix $\lambda_1$ to the unit vector along the first latent coordinate, anchoring its scale and direction \citep{bollen1989structural}. The strictly lower-triangular matrix $\Psi$ describes dependence on earlier entries in the same behavior vector. Shared propensity and within-arrival dependence both contribute to behavioral co-occurrence.

Let $\E_{P_{g,t}}$ denote expectation over the current propensity distribution. Integrating the contact--behavior product gives
\begin{equation}
\begin{aligned}
\widehat N_{jgt}&=n_{jgt}\E_{P_{g,t}}[\ell_{jgt}(U)],\\
q_{jgt}(y)&=\frac{\E_{P_{g,t}}[\ell_{jgt}(U)p_\theta(y\mid U,x_{jgt},s_t)]}{\E_{P_{g,t}}[\ell_{jgt}(U)]},\\
\widehat Y_{jgth}&=\widehat N_{jgt}\sum_y y_hq_{jgt}(y).
\end{aligned}
\label{eq:observations}
\end{equation}
The numerator is the intensity of pattern $y$, and the denominator is the intensity of all arrivals. Their ratio is the probability that a recorded arrival exhibits pattern $y$. A small population component can therefore contribute many events if it is frequently reached \citep{patil1978weighted}. Multiplication before integration preserves the association between contact intensity and response probability.

At the reparameterized nodes, the behavioral probability is computed as
\begin{equation}
\widehat q_{jgt}(y)=\frac{\sum_{k,i}\pi_{gk}w_i\ell_{jgt}(u_{gkit})p_\theta(y\mid u_{gkit},x_{jgt},s_t)}{\sum_{k,i}\pi_{gk}w_i\ell_{jgt}(u_{gkit})}.
\label{eq:quadrature}
\end{equation}
Section~\ref{sec:learning} uses this approximation and writes it as $q$. The same weighted sums produce arrivals and behavioral totals. We enumerate all $2^H$ patterns; positive weights preserve normalization. At zero effort, expected counts are zero and conditional probabilities use the positive-effort limit. The arrival and behavior parameters together form $\theta_O$.

\subsection{Feedback Recurrence}
\label{sec:recurrence}

The state $s_t$ contains response shifts $\delta_{g,t}$, behavioral memory $m_{g,t}$, fatigue $f_{g,t}$, and arrival states $z_t,c_{jt}$. Let $\widetilde N_{jgt}$ be the arrival count and $\widetilde Y_{jgt}=(\widetilde Y_{jgth})_{h=1}^{H}$ the vector of behavioral counts used for feedback. They contain recorded outcomes during observation and model expectations during forecasting. Cohort feedback comprises a smoothed behavioral-rate vector $v_{g,t}$ and contact dose $o_{g,t}$:
\begin{equation}
v_{g,t}=\frac{\sum_j\widetilde Y_{jgt}+2v_g^0}{\sum_j\widetilde N_{jgt}+2},
\qquad o_{g,t}=\frac{\sum_j\widetilde N_{jgt}}{\max(1,n_g^{\mathrm{ref}})},
\label{eq:feedback}
\end{equation}
where $v_g^0$ is the reference behavioral-rate vector and $n_g^{\mathrm{ref}}$ is the maximum cohort reference size across units. Feedback updates the response state through
\begin{equation}
\begin{aligned}
m_{g,t+1}&=\rho_m m_{g,t}+(1-\rho_m)v_{g,t},\\
f_{g,t+1}&=\rho_f f_{g,t}+(1-\rho_f)(1-e^{-o_{g,t}}),\\
\delta_{g,t+1}&=\rho_\delta\delta_{g,t}+B^{\top}(v_{g,t}-v_g^0)-f_{g,t+1}b_f.
\end{aligned}
\label{eq:transition}
\end{equation}
The learned retention factors $\rho_m,\rho_f,\rho_\delta\in(0,1)$ control persistence through geometric carryover, a common form in marketing-response models \citep{hanssens2001market,jin2017bayesian}; separate factors allow memory and fatigue to persist for different durations \citep{kronrod2019ad}. The fatigue input $1-e^{-o_{g,t}}$ saturates as contact dose grows, reflecting the diminishing marginal effect of repeated exposure \citep{krugman1972why,pechmann1988advertising}. The matrix $B$ maps behavioral-rate deviations to propensity shifts, and $b_f$ maps fatigue to those shifts. Memory and fatigue also enter the behavioral logits directly. Current experience thus affects both the distribution and responses at a given propensity.

Arrival discrepancies update the common environment:
\begin{equation}
\xi_{jt}=\log(1+\sum_g\widetilde N_{jgt})-\log(1+\sum_g\widehat N_{jgt}),
\qquad z_{t+1}=\rho_z z_t+\kappa_z\bar\xi_t,
\label{eq:innovation}
\end{equation}
where $\xi_{jt}$ is the unit's log-count innovation, $\bar\xi_t$ is its mean across active units, and $\rho_z,\kappa_z$ are learned retention and correction coefficients. This correction follows the innovations form of exponential-smoothing state-space models \citep{hyndman2008forecasting}. Unit states receive centered innovations and are centered to sum to zero (Appendix~\ref{app:observation}). All transition parameters form $\theta_T$. Forecasts fix historical summaries and reference sizes at the origin, advance known calendar and effort inputs, and feed predicted counts into the recurrence. Expected feedback gives zero arrival innovation while memory, fatigue, and response shifts continue evolving.

\subsection{Joint Learning from Aggregate Counts}
\label{sec:learning}

DGD learns $\theta=(\theta_P,\theta_O,\theta_T)$ through the complete prediction--feedback sequence. Arrival counts follow a negative-binomial working likelihood $\NB(N;\widehat N,r)$ with mean $\widehat N$, dispersion $r$, and variance $\widehat N+\widehat N^2/r$ \citep{cameron2013regression}. Conditional on arrivals, pattern frequencies follow a multinomial likelihood with probabilities $q$. For daily cells $\mathcal B_t$, define
\begin{equation}
\mathcal L_{B,t}=-\frac{\sum_{(j,g)\in\mathcal B_t}\sum_y C_{jgt}(y)\log q_{jgt}(y)}{\max(1,\sum_{(j,g)\in\mathcal B_t}N_{jgt})},\quad
\mathcal L_{N,t}=-\frac{\sum_{(j,g)\in\mathcal B_t}\log\NB(N_{jgt};\widehat N_{jgt},r)}{|\mathcal B_t|}.
\label{eq:losses}
\end{equation}
The parameter-independent multinomial constant is omitted. Behavioral losses are normalized by daily event counts and count losses by daily cell counts.

The observed branch advances the state with recorded counts, as in teacher forcing \citep{williams1989learning}, while forecasts advance it with predicted counts. To train the updates used in forecasting, an \emph{expected-feedback branch} starts from the same $s_t$ as the observed branch, uses predicted day-$t$ counts to construct $s_{t+1\mid t}^{\mathrm{free}}$, and scores the next day's labels under the forecast input protocol in Section~\ref{sec:recurrence}. Scheduled sampling and multi-step time-series training likewise feed a model's own predictions back during training \citep{bengio2015scheduled,venkatraman2015improving}. The branch loss is $\mathcal L_{t+1\mid t}^{\mathrm{free}}=\mathcal L_{B,t+1\mid t}^{\mathrm{free}}+\alpha\mathcal L_{N,t+1\mid t}^{\mathrm{free}}$. The objective over $T$ training days is
\begin{equation}
\mathcal L(\theta)=\frac1T\left[\sum_{t=1}^T(\mathcal L_{B,t}+\alpha\mathcal L_{N,t})+\beta\sum_{t=1}^{T-1}\mathcal L_{t+1\mid t}^{\mathrm{free}}\right]+\lambda_R\mathcal R(\theta),
\label{eq:objective}
\end{equation}
where $\alpha,\beta,\lambda_R$ weight the count, forecast, and regularization terms. The parameter-normalized regularizer $\mathcal R$ penalizes main coefficients and departures of Gaussian scales from initialization. The computation graph is retained across the sequence, so gradients propagate backward through time \citep{werbos1990backpropagation}. Current losses train mixture weights and node locations through integration; later losses additionally train retention and feedback through recurrence. Algorithm~\ref{alg:training} in Appendix~\ref{app:training} gives the complete procedure.

\paragraph{Properties of aggregate learning.}
For conditionally independent events sharing context and probability vector $q$, training from daily pattern counts yields the same parameter gradients as the event likelihood: their log-likelihoods differ only by a parameter-independent combinatorial term. Under unclipped exponential-linear intensity, arrival weighting shifts component means and reweights components, preserving a Gaussian mixture for recorded arrivals. With a shared arrival loading and common component shift, these changes can be absorbed into mixture parameters and the arrival intercept, yielding an equivalent separated readout. Expected feedback propagates finite-quadrature error to later states and predictions. Appendices~\ref{app:equivalence}--\ref{app:numerical} derive these properties and probability, gradient, and state-error bounds; Appendix~\ref{sec:numerical-results} checks them numerically.

%% file: sections/04-experiments.tex
\section{Experiments}
\label{sec:experiments}

We evaluate DGD's collective-response predictions, then examine how learning its population representation and feedback recurrence affects those predictions. Controlled experiments test response recovery under known dynamics and new product contexts.

\subsection{Experimental Setup}
\label{sec:setup}

\paragraph{Model instance.}
In all four real-data windows, DGD is a probabilistic model trained in PyTorch \citep{paszke2019pytorch} with 64-bit arithmetic. Each cohort has \textbf{two one-dimensional Gaussian components}, integrated with \textbf{seven Gauss--Hermite nodes per component}. The clipped exponential-linear arrival function and autoregressive logistic behavior function share the latent propensity. DGD enumerates 128 joint patterns for KuaiRand's seven behaviors and eight patterns for Retail's three basket marks. It learns distribution, observation, and transition parameters together; arrival counts use a negative-binomial likelihood with dispersion 50. Each window is fitted independently with three training seeds.

\paragraph{Data and prediction targets.}
KuaiRand-Pure provides standard-recommendation and random-exposure logs \citep{gao2022kuairand}. Both windows use nine cohorts fixed from preceding history and five observation units: four creators and a pooled unit. Each exposure records clicking, long viewing, liking, following, commenting, forwarding, and negative feedback. Online Retail II provides two annual invoice windows \citep{chen2012onlineretailii}. Each uses four customer cohorts and one store; the binary marks indicate merchandise value, quantity, and distinct-product count above warm-up medians. The three known-customer cohorts are single-purchase, low-spending repeat, and high-spending repeat customers; a fourth contains customers absent from warm-up or without identifiers. Features contain calendar inputs and preceding cohort summaries, and activity effort is $e=1$. All windows use chronological train/validation/test splits (Table~\ref{tab:data}); preprocessing and dates are in Appendix~\ref{app:protocol}.

\paragraph{Baselines and fitting.}
DeepAR-joint adapts DeepAR's shared LSTM and negative-binomial count output with an autoregressive binary-mark head \citep{salinas2020deepar}. KuaiRand also includes two GRU predictors \citep{cho2014learning}: a full-projection GRU learns every behavior's hidden-state projection, while a restricted-projection GRU fixes the first behavior's projection to the first hidden coordinate. Neural comparisons share the available aggregate inputs. Three historical baselines use training-period, last-day, or exponentially smoothed counts and joint distributions. DGD uses Adam \citep{kingma2015adam} with learning rate 0.025, gradient-norm clipping at 10 \citep{pascanu2013difficulty}, and loss weights $\alpha=0.05$, $\beta=0.15$, $\lambda_R=0.001$. Epoch budgets are 300 for KuaiRand and 120 for Retail, with patience 20. Validation selects epochs, DeepAR widths from 8 and 16, and historical smoothing settings. Appendix~\ref{app:protocol} supplies the complete protocols.

\paragraph{Metrics.}
Joint behavioral NLL averages the negative log-likelihood of each recorded event's joint behavior pattern over all test events. Mean Brier score averages squared errors between marginal probabilities and binary outcomes, equally across behaviors. Cohort-rate MAE weights cohorts equally and is reported in percentage points (pp). Behavioral-count MAE measures daily count errors averaged over behaviors or basket marks. Forecasts combine each model's own predicted arrivals and behavioral probabilities. Neural scores average three per-seed metrics; reported standard deviations describe training-seed variation. Appendix~\ref{app:behavior-scoring} gives scoring formulas and forecast settings.

\subsection{Collective Response Prediction}
\label{sec:prediction-results}

\textbf{DGD improves joint, marginal, and cohort-level behavioral prediction.}
Table~\ref{tab:behavior} compares the four windows. DGD's joint NLL is lower than DeepAR-joint in every window and lower than both GRUs in the two KuaiRand windows. Under standard recommendation, NLL is 1.1232 versus 1.3466 for DeepAR-joint, while cohort-rate MAE is 4.21 versus 4.90 pp (Table~\ref{tab:response}). DGD also has lower Brier scores and cohort-rate errors than these recurrent comparators in each evaluated window. The improvements span the joint patterns used in training, individual behaviors, and equally weighted cohort responses. Section~\ref{sec:heterogeneity-results} examines the contribution of distribution learning.

\input{tables/behavior-nll.tex}
\input{tables/behavior-response.tex}

\textbf{DGD also improves aggregate response-count prediction.}
In Retail, DGD has lower one-day behavioral-count MAE than every listed baseline in both annual windows (Table~\ref{tab:retail-counts}). Errors are 4.71 and 5.38 invoices/day, compared with DeepAR-joint's 6.88 and 5.77. The reductions relative to the lowest-error historical baselines are 54.60\% and 54.48\%, respectively. These predictions combine event volume and conditional behavior: the shared Gaussian representation generates arrivals and the probability that an invoice bears each basket mark. Appendix~\ref{app:trajectories} plots the daily forecasts.

\input{tables/retail-behavior-counts.tex}

\subsection{Learning the Population Representation}
\label{sec:heterogeneity-results}

The representation ablation retains the same observation functions, feedback structure, and aggregate histories. Fixed Gaussian freezes mixture weights, base means, and covariances; readout and transition parameters remain trainable. The matched discrete model uses three trainable support points, giving the same five effective distribution parameters per cohort as DGD; their initial means and total variances are matched. The point representation removes within-cohort heterogeneity.

Learning DGD's distribution reduces standard-recommendation NLL from 1.2594 to 1.1232, a \textbf{10.82\% reduction} (Table~\ref{tab:representation}). Brier, cohort-rate, and pairwise co-occurrence errors also decrease. Thus, gradients from aggregate observations improve the representation used to predict individual behaviors and their combinations. The point model has NLL 1.2443 and cohort MAE 5.74 pp. Gaussian and matched discrete mixtures both improve these predictions, attaining NLL 1.1232 and 1.1339, respectively.

\input{tables/ablation-representation.tex}

\subsection{Learning Feedback for Future Responses}
\label{sec:dynamics-results}

The internal-dynamics ablation removes trainable recurrence while retaining observed lag summaries. Under random exposure, recurrence reduces one-day behavioral-count MAE from 34.08 to 16.47 and three-day MAE from 38.32 to 22.76 (Table~\ref{tab:dynamic-ablation}). Because both variants retain the same observed summaries, the comparison evaluates the contribution of propagating current responses into subsequent state.

\input{tables/ablation-dynamics.tex}

The controlled dynamics experiment fits the same arrival--response model to counts simulated from a generator with behavioral memory and response shifts. It uses two cohorts, two units, two behaviors, and two one-dimensional Gaussian components, with nine nodes per component and negative-binomial dispersion 80. The split contains 20 training, 8 validation, and 12 test days. DGD attains mean joint KL 0.0340 versus 0.2577 without dynamics (Table~\ref{tab:dynamics}), with lower joint KL and marginal probability error in all three seeds. Sequence losses passing through recurrence recover the generator's cross-day response distributions from cohort counts.

\input{tables/dynamics-recovery.tex}

\subsection{Response Prediction under New Attributes and Prices}
\label{sec:forecast-results}

This experiment evaluates responses to new product attributes and prices under known Gaussian and discrete generators. The controlled choice task independently fits a \textbf{two-component, two-dimensional Gaussian mixture} and a softmax readout over three products and an outside option. The readout models attribute preferences and positive price sensitivity; its integrated probabilities predict market-level purchase counts (Appendix~\ref{app:observation}). Gaussian integration uses seven nodes per dimension. Each seed contains 20 training, 8 validation, and 40 test markets, each with 4,000 opportunities. The test markets are held out from fitting and validation.

Relative to a fixed Gaussian, learning the distribution reduces choice KL from 0.001994 to 0.001207 under the Gaussian generator and from 0.020016 to 0.002542 under the discrete generator (Table~\ref{tab:choice}). The point representation has larger errors under both generators. These results show that mixture responses learned from aggregate purchases predict choice distributions in new product contexts.

\input{tables/choice-recovery.tex}

%% file: tables/behavior-nll.tex
\begin{table}[!htbp]
  \centering
  \caption{Joint behavioral NLL of DGD and recurrent baselines (lower is better). Values are means $\pm$ training-seed standard deviations; comparisons are within columns. DeepAR-joint adds a joint-mark head to DeepAR. GRUs were evaluated on KuaiRand; dashes denote unevaluated settings.}
  \label{tab:behavior}
  \footnotesize
  \setlength{\tabcolsep}{3pt}
  \renewcommand{\arraystretch}{1.12}
  \begin{tabular}{@{}lrrrr@{}}
    \toprule
    Model & Standard & Random & Retail 2010 & Retail 2011 \\
    \midrule
    DGD & $1.1232 \pm 0.0005$ & $0.7591 \pm 0.0762$ & $1.7357 \pm 0.0012$ & $1.7555 \pm 0.0008$ \\
    DeepAR-joint (adapted) & $1.3466 \pm 0.0224$ & $0.7899 \pm 0.0252$ & $1.8819 \pm 0.0205$ & $1.8305 \pm 0.0030$ \\
    Full-projection GRU & $1.2290 \pm 0.0076$ & $0.7970 \pm 0.0094$ & --- & --- \\
    Restricted-projection GRU & $1.2307 \pm 0.0202$ & $0.7972 \pm 0.0084$ & --- & --- \\
    \bottomrule
  \end{tabular}
\end{table}

%% file: tables/behavior-response.tex
\begin{table}[!htbp]
  \centering
  \caption{Marginal and cohort-level predictions compared with recurrent baselines. Each cell gives mean Brier / cohort-rate MAE in percentage points (both lower is better). Neural scores average three training seeds. Appendix~\ref{app:behavior-scoring} defines the scores.}
  \label{tab:response}
  \small
  \setlength{\tabcolsep}{4pt}
  \renewcommand{\arraystretch}{1.12}
  \begin{tabular}{@{}lrrrr@{}}
    \toprule
    Model & Standard & Random & Retail 2010 & Retail 2011 \\
    \midrule
    DGD & $0.0683 / 4.21$ & $0.0324 / 1.36$ & $0.2482 / 12.84$ & $0.2396 / 14.82$ \\
    DeepAR-joint (adapted) & $0.0709 / 4.90$ & $0.0335 / 1.86$ & $0.2535 / 14.26$ & $0.2425 / 15.52$ \\
    Full-projection GRU & $0.0726 / 5.56$ & $0.0325 / 1.39$ & --- & --- \\
    Restricted-projection GRU & $0.0731 / 5.62$ & $0.0325 / 1.40$ & --- & --- \\
    \bottomrule
  \end{tabular}
\end{table}

%% file: tables/retail-behavior-counts.tex
\begin{table}[!htbp]
  \centering
  \caption{One-day behavioral-count MAE in Retail (invoices/day; lower is better). Absolute daily errors are computed separately for high-value, high-quantity, and many-SKU marks, then averaged over dates and marks. Each window has 17 origins, with identical origin--endpoint pairs across methods. Relative improvements use unrounded errors.}
  \label{tab:retail-counts}
  \small
  \setlength{\tabcolsep}{12pt}
  \renewcommand{\arraystretch}{1.12}
  \begin{tabular}{@{}lrr@{}}
    \toprule
    Model & Retail 2010 & Retail 2011 \\
    \midrule
    DGD & $4.71$ & $5.38$ \\
    DeepAR-joint (adapted) & $6.88$ & $5.77$ \\
    Training history & $13.53$ & $14.34$ \\
    Last-day history & $12.36$ & $11.83$ \\
    Smoothed history & $10.38$ & $12.29$ \\
    \bottomrule
  \end{tabular}
\end{table}

%% file: tables/ablation-representation.tex
\begin{table}[!htbp]
  \centering
  \caption{Representation and distribution-learning ablations under standard recommendation. Values are means $\pm$ training-seed standard deviations (lower is better). Rate errors are in percentage points (pp); co-occurrence error covers all 21 pairs of the seven behaviors.}
  \label{tab:representation}
  \footnotesize
  \setlength{\tabcolsep}{3pt}
  \renewcommand{\arraystretch}{1.12}
  \begin{tabular}{@{}lrrrr@{}}
    \toprule
    Model & Joint NLL & Mean Brier & Cohort MAE (pp) & Pair MAE (pp) \\
    \midrule
    DGD & $1.1232 \pm 0.0005$ & $0.0683 \pm 0.0001$ & $4.21 \pm 0.02$ & $0.393 \pm 0.016$ \\
    Fixed Gaussian & $1.2594 \pm 0.0127$ & $0.0761 \pm 0.0006$ & $6.36 \pm 0.20$ & $0.890 \pm 0.115$ \\
    Matched discrete & $1.1339 \pm 0.0074$ & $0.0685 \pm 0.0003$ & $4.25 \pm 0.10$ & $0.404 \pm 0.028$ \\
    Point representation & $1.2443 \pm 0.0027$ & $0.0734 \pm 0.0002$ & $5.74 \pm 0.07$ & $0.962 \pm 0.020$ \\
    \bottomrule
  \end{tabular}
\end{table}

%% file: tables/ablation-dynamics.tex
\begin{table}[!htbp]
  \centering
  \caption{Internal-dynamics ablation under random exposure. Values are means $\pm$ training-seed standard deviations of behavioral-count MAE, averaged over seven behaviors (occurrences/day). One-day and three-day forecasts have seven and five eligible origins, respectively.}
  \label{tab:dynamic-ablation}
  \small
  \setlength{\tabcolsep}{12pt}
  \renewcommand{\arraystretch}{1.12}
  \begin{tabular}{@{}lrr@{}}
    \toprule
    Model & One-day MAE & Three-day MAE \\
    \midrule
    DGD & $16.47 \pm 0.85$ & $22.76 \pm 0.50$ \\
    Without internal dynamics & $34.08 \pm 0.59$ & $38.32 \pm 0.52$ \\
    \bottomrule
  \end{tabular}
\end{table}

%% file: tables/dynamics-recovery.tex
\begin{table}[!htbp]
  \centering
  \caption{Response recovery under a known dynamic generator. KL compares predicted and true joint distributions; marginal MAE compares behavioral probabilities. Values are means $\pm$ seed standard deviations.}
  \label{tab:dynamics}
  \small
  \setlength{\tabcolsep}{4pt}
  \renewcommand{\arraystretch}{1.12}
  \begin{tabular}{@{}lrr@{}}
    \toprule
    Model & Joint KL & Marginal MAE \\
    \midrule
    DGD & $0.0340 \pm 0.0299$ & $0.0519 \pm 0.0337$ \\
    Without dynamics & $0.2577 \pm 0.0052$ & $0.1661 \pm 0.0042$ \\
    \bottomrule
  \end{tabular}
\end{table}

%% file: tables/choice-recovery.tex
\begin{table}[!htbp]
  \centering
  \caption{Choice-distribution recovery under known market generators. Values are the seed mean $\pm$ standard deviation of average KL across test markets.}
  \label{tab:choice}
  \small
  \setlength{\tabcolsep}{4pt}
  \renewcommand{\arraystretch}{1.12}
  \begin{tabular}{@{}lrr@{}}
    \toprule
    Model & Gaussian generator & Discrete generator \\
    \midrule
    Learned Gaussian & $0.001207 \pm 0.000544$ & $0.002542 \pm 0.002337$ \\
    Fixed Gaussian & $0.001994 \pm 0.000522$ & $0.020016 \pm 0.027627$ \\
    Point representation & $0.059223 \pm 0.005277$ & $0.063599 \pm 0.010622$ \\
    \bottomrule
  \end{tabular}
\end{table}

%% file: sections/05-conclusion.tex
\section{Conclusion}
\label{sec:conclusion}

DGD jointly learns a Gaussian response representation, contact--behavior aggregation, and feedback recurrence from aggregate counts. Across four recommendation and retail windows, it improves joint behavioral prediction over DeepAR-joint; ablations establish the contributions of distribution learning and feedback. Controlled experiments recover cross-day dynamics and responses under new attributes and prices.

%% file: appendix.tex
\section{Model Computation Details}
\label{app:observation}

The Gaussian mixture represents variation in latent response propensities within a cohort. The observation readout converts these propensities into response probabilities and event volumes under a given context. Social-media exposures and retail invoices use arrival and joint binary-behavior readouts. The controlled choice experiment pairs the same form of distributional representation with a choice-probability readout and fits it independently from market counts.

\subsection{Aggregate Observation Likelihoods}

The arrival likelihood describes the total number of events generated by a cohort within an observation unit; the behavioral likelihood describes the composition of their joint responses. Using the notation of the main text, the negative-binomial distribution has mean $\widehat N$, dispersion $r$, and log-probability
\begin{equation}
\begin{aligned}
\log p(N\mid\widehat N,r)
={}&\log\Gamma(N+r)-\log\Gamma(N+1)-\log\Gamma(r)\\
&+r\log\frac{r}{r+\widehat N}+N\log\frac{\widehat N}{r+\widehat N}.
\end{aligned}
\end{equation}
Real-data training fixes $r=50$ and sets the mean to the predicted arrival count. Given the arrival count $N$, behavioral-pattern counts $C$ follow a multinomial distribution with joint probabilities $q$:
\begin{equation}
p(C\mid N,q)=\frac{N!}{\prod_y C(y)!}\prod_y q(y)^{C(y)}.
\end{equation}
Within each date, the behavioral loss is divided by the number of events, with a minimum denominator of one, and the count loss is divided by the number of unit--cohort cells. The losses are then averaged over dates.

The expected-feedback forecast term starts from the current day's initial state, takes one expected-feedback step to obtain the next day's state, and evaluates behavioral and weighted count losses against the next day's aggregate labels. The validation objective contains only daily behavioral and weighted count losses. Test behavioral NLL is normalized by the total number of test events, using the event-weighted evaluation in the main text.

\subsection{State Updates}

The average prediction discrepancy across active units corrects the common arrival environment; each unit's departure from that average corrects its offset. Let $\mathbf{1}_{jt}$ indicate whether the sum, over cohorts, of cohort reference size times activity effort is positive for unit $j$ on day $t$. Using the log innovations in Equation~\eqref{eq:innovation}, the mean innovation across active units is
\begin{equation}
\bar\xi_t=\frac{\sum_j \mathbf{1}_{jt}\xi_{jt}}{\max(1,\sum_j \mathbf{1}_{jt})}.
\end{equation}
Each unit's state retains part of its previous offset and is corrected according to the difference between its innovation and the mean innovation. The updated states are then centered across units:
\begin{equation}
\begin{aligned}
\widetilde c_{j,t+1}&=\rho_zc_{jt}
+\kappa_c \mathbf{1}_{jt}(\xi_{jt}-\bar\xi_t),\\
c_{j,t+1}&=\widetilde c_{j,t+1}
-\frac1J\sum_{j'}\widetilde c_{j',t+1}.
\end{aligned}
\end{equation}
Centering keeps the sum of unit-specific states at zero, so these states represent offsets relative to the common arrival environment. The common environment follows Equation~\eqref{eq:innovation}. Retention factors $\rho_\delta,\rho_z,\rho_f,\rho_m$ and correction gains $\kappa_z,\kappa_c$ are all sigmoid-parameterized. Initial response shifts, memory, fatigue, and arrival states are zero; warm-up statistics set baseline intensities and reference behavioral rates.

During forecasting, model expectations provide feedback. With $\widetilde N=\widehat N$, the arrival innovation $\xi_{jt}$ is zero, and common and unit-specific states decay with factor $\rho_z$. Behavioral memory, fatigue, and response shifts continue to follow Equations~\eqref{eq:feedback} and~\eqref{eq:transition}, advancing the cohort's response state to later dates.

\subsection{Choice Response Readout}

The controlled choice experiment learns heterogeneous consumer responses to product attributes and prices from market-level purchase counts, following random-coefficient logit demand models \citep{berry1995automobile,mcfadden2000mixed}. Here, latent propensity $u$ modulates attribute preferences and price sensitivity, and the choice readout converts it into probabilities of purchasing each product or making no purchase. Each product $b$ has attributes $x_b$, price $\mathrm{price}_b$, and an availability indicator. Conditional on $u$, its utility is
\begin{equation}
V_b(u)=x_b^{\top}\beta_C+u^{\top}W_C^{\top}x_b
-\mathrm{price}_b\softplus(a_C+v_C^{\top}u),
\end{equation}
where $\beta_C$ gives baseline attribute preferences, $W_C$ maps latent propensity to attribute preferences, and $a_C,v_C$ determine the price-sensitivity intercept and loading. These are trainable choice-readout parameters. The softplus parameterization keeps price sensitivity positive. The outside option represents making no purchase on the current occasion and has utility zero. For the available product set $\mathcal A$, the softmax denominator includes only products in that set:
\begin{equation}
\begin{aligned}
p(b\mid u)&=\frac{\exp V_b(u)}{1+\sum_{b'\in\mathcal A}\exp V_{b'}(u)},\quad b\in\mathcal A,\\
p(0\mid u)&=\frac{1}{1+\sum_{b'\in\mathcal A}\exp V_{b'}(u)}.
\end{aligned}
\end{equation}
Integrating choice probabilities over each cohort's propensity distribution, weighting by cohort reference size, and summing across cohorts gives aggregate demand $Q_b=\sum_g n_g\E_{P_g}[p(b\mid U)]$. Including the outside option yields $\sum_{b\in\mathcal A\cup\{0\}}Q_b=\sum_gn_g$, allocating all choice occasions to available products or no purchase. The choice experiment samples market counts from a multinomial distribution parameterized by aggregate probabilities and independently fits distribution and choice-readout parameters with the corresponding count objective.

\section{Equivalence under Aggregate Observations}
\label{app:equivalence}

\subsection{Likelihood Sufficiency of Counts under Shared Conditions}

Counts retain the parameter information in the event likelihood when the pooled events share the same conditional distribution. Suppose $N$ events are conditionally independent given state and context and share joint probability $q(y)$. Collecting terms with the same behavioral pattern gives
\begin{equation}
\sum_{\nu=1}^{N}\log q(y_\nu)=\sum_y C(y)\log q(y).
\end{equation}
There are $N!/\prod_y C(y)!$ event orderings with the same count vector, each with equal probability under the shared conditional distribution. Summing their probabilities adds a combinatorial constant to the log-likelihood above. This constant is independent of the parameters, so event and count objectives have identical parameter gradients. Collective graphical models likewise treat such count vectors as sufficient statistics of individual models for inference \citep{sheldon2011collective}. Conditional independence applies across events; conditional dependence among behaviors within an event remains represented by $q(y)$.

\subsection{Exponential Tilting of Gaussian Mixtures}

Exponential arrival intensity changes the contribution of each response type to the observations, producing a weighted distribution of propensities \citep{patil1978weighted}. For the Gaussian mixture representing response heterogeneity, this reweighting can be expressed through a transformed set of mixture parameters. Fix a cohort and suppress its index. Let $\phi(u;m,\Sigma)$ denote the Gaussian density with mean $m$ and covariance $\Sigma$. Completing the square in its exponent yields
\begin{equation}
e^{\gamma^{\top}u}\phi(u;m,\Sigma)
=\exp\!\left(\gamma^{\top}m+\tfrac12\gamma^{\top}\Sigma\gamma\right)
\phi(u;m+\Sigma\gamma,\Sigma).
\end{equation}
Substitute $m=\mu_k+\delta_t$ and define
\begin{equation}
\zeta_k=\exp\!\left(\gamma^{\top}\mu_k+\tfrac12\gamma^{\top}\Sigma_k\gamma\right),
\qquad Z=\sum_k\pi_k\zeta_k.
\end{equation}
The cohort's dynamic shift contributes the same factor $\exp(\gamma^{\top}\delta_t)$ to every component, so this factor cancels after normalization. The original component weights are reweighted by $\zeta_k$, and the component means shift in the direction determined by their covariances and the arrival loading. The propensity distribution reweighted by exponential intensity is therefore
\begin{equation}
\begin{aligned}
P_E&=\sum_k\widetilde\pi_k
\mathcal N(\widetilde\mu_k+\delta_t,\Sigma_k),\\
\widetilde\pi_k&=\frac{\pi_k\zeta_k}{Z},\qquad
\widetilde\mu_k=\mu_k+\Sigma_k\gamma.
\end{aligned}
\end{equation}
Set $\widetilde P=P_E$ as the response-propensity distribution of a separated-readout model. Directly integrating the same behavioral function gives $\widetilde q(y)=\E_{\widetilde P}[p(y\mid U)]=q(y)$. The observation weights assigned to response types by the original arrival intensity are now part of the transformed propensity distribution. Preserving arrival totals also requires an intercept adjustment. Define
\begin{equation}
\begin{aligned}
\widetilde Z&=\sum_k\widetilde\pi_k
\exp\!\left(\gamma^{\top}\widetilde\mu_k+
\tfrac12\gamma^{\top}\Sigma_k\gamma\right),\\
\widetilde a&=a+\log Z-\log\widetilde Z.
\end{aligned}
\end{equation}
The arrival part of the separated model retains the same arrival loading and uses the new intercept. The intercept adjustment cancels the change in integrated intensity caused by transforming the propensity distribution; the behavioral part uses the direct integral above. Both readouts therefore give the same arrival totals, joint behavioral probabilities, and behavioral totals. Identifiability results for finite Gaussian mixtures concern observed mixture densities \citep{teicher1963finite,yakowitz1968identifiability}; this equivalence concerns integrals of the arrival and behavioral readouts.

If the models start from the same auxiliary states, deterministic observed or expected feedback maps their identical daily outputs to the same next-day states. Induction over days establishes path equivalence for any finite horizon.

This transformation applies to unclipped exponential intensity, an arrival loading shared across contexts, a dynamic shift shared by mixture components, and adjustable intercepts. The implemented quadrature model also includes intensity clipping and regularization. Section~\ref{sec:numerical-results} evaluates this transformation numerically with fixed fitted parameters.

\section{Numerical Analysis of DGD}
\label{app:numerical}

\subsection{Probabilistic Properties of Positive-Weight Quadrature}

Finite quadrature expresses the population response as a weighted sum of responses at latent nodes. Let $\widetilde w_{ki}=\pi_kw_i\ell(u_{ki})\geq0$ and assume $\sum_{ki}\widetilde w_{ki}>0$. At each node, the autoregressive joint distribution sums to one across behavioral patterns. The normalized population probability therefore satisfies
\begin{equation}
\sum_y\widehat q(y)
=\frac{\sum_{ki}\widetilde w_{ki}\sum_yp(y\mid u_{ki})}{\sum_{ki}\widetilde w_{ki}}=1.
\end{equation}
Expected behavioral counts are also nonnegative weighted sums of node-level expectations. Summing first within each Gaussian component and then across components only reorders a finite sum, so the component contributions sum to the population total. Each node's probability of a binary behavior lies between zero and one, giving $0\leq\widehat Y_h\leq\widehat N$.

\subsection{Quadrature Error Bounds}

The joint behavioral probability is the ratio of behavior-weighted intensity to total arrival intensity, so quadrature error affects both its numerator and denominator. Let $q=A/D$ and $\widehat q=\widehat A/\widehat D$ be valid probabilities. Assume
\begin{equation}
D\geq d_0>0,\quad
|\widehat D-D|\leq\varepsilon_D<d_0,\quad
|\widehat A-A|\leq\varepsilon_A.
\end{equation}
The identity
\begin{equation}
\widehat q-q=
\frac{(\widehat A-A)-q(\widehat D-D)}{\widehat D}
\end{equation}
together with $0\leq q\leq1$ bounds the numerator by the sum of the two integration errors and the denominator below by $d_0-\varepsilon_D$, giving
\begin{equation}
|\widehat q-q|\leq\frac{\varepsilon_A+\varepsilon_D}{d_0-\varepsilon_D}.
\label{eq:quadrature-bound}
\end{equation}
To examine the effect on training gradients, further assume that derivatives of the integrals exist and satisfy
\begin{equation}
\begin{aligned}
\|\nabla\widehat A-\nabla A\|&\leq\varepsilon_{\nabla A},&
\|\nabla\widehat D-\nabla D\|&\leq\varepsilon_{\nabla D},\\
\|\nabla D\|&\leq G_D,&\|\nabla q\|&\leq G_q.
\end{aligned}
\end{equation}
Write $\Delta_q=|\widehat q-q|$ and apply the quotient rule to the exact and approximate probabilities. Expand their gradient difference into errors in the numerator derivative, denominator derivative, probability value, and denominator value. The triangle inequality then gives
\begin{equation}
\|\nabla\widehat q-\nabla q\|
\leq
\frac{\varepsilon_{\nabla A}+\varepsilon_{\nabla D}
+\Delta_qG_D+G_q\varepsilon_D}{d_0-\varepsilon_D}.
\end{equation}
If $\min(q,\widehat q)\geq q_{\min}>0$, then
\begin{equation}
\begin{aligned}
|\log\widehat q-\log q|&\leq\Delta_q/q_{\min},\\
\|\nabla\log\widehat q-\nabla\log q\|
&\leq\frac{\|\nabla\widehat q-\nabla q\|}{q_{\min}}
+\frac{G_q\Delta_q}{q_{\min}^2}.
\end{aligned}
\end{equation}
Weighting by empirical behavioral frequencies gives error bounds for average behavioral NLL and its gradient. Under reparameterized training, both mixture weights and latent node locations vary with the parameters, and their derivatives contribute to the integral's gradient.

\subsection{Error Propagation and State Bounds under Expected Feedback}

Expected feedback carries observation-integration errors into subsequent states. The recurrence error therefore includes both the approximation made in the current update and the propagation of existing state errors. Let exact and approximate expected-feedback updates be $s_{t+1}=\Phi_t(s_t)$ and $\widehat s_{t+1}=\widehat\Phi_t(\widehat s_t)$. Suppose $\Phi_t$ has Lipschitz constant $L_t$ in the relevant trajectory region and $\sup_s\|\widehat\Phi_t(s)-\Phi_t(s)\|\leq\varepsilon_t$. Then
\begin{equation}
\begin{aligned}
E_{t+1}&\leq L_tE_t+\varepsilon_t,\\
E_T&\leq E_0\prod_{t=0}^{T-1}L_t+
\sum_{j=0}^{T-1}\varepsilon_j\prod_{t=j+1}^{T-1}L_t,
\end{aligned}
\end{equation}
where $E_t=\|\widehat s_t-s_t\|$. To obtain the one-step bound, add and subtract $\Phi_t(\widehat s_t)$ in the state difference. This separates the approximation error at the same input state, bounded by $\varepsilon_t$, from the exact update's response to the two input states, bounded by $L_tE_t$. Repeated substitution gives the finite-horizon sum. If $L_t\leq L<1$ and $\varepsilon_t\leq\varepsilon$, then $E_T\leq L^TE_0+\varepsilon(1-L^T)/(1-L)$.

For fixed finite model parameters, suppose initial memory, fatigue, and reference behavioral rates lie in $[0,1]$, and effort and cohort reference sizes are nonnegative. The behavioral-count bound in Appendix~C.1 places smoothed behavioral rates under expected feedback in $[0,1]$. Nonnegative dose also places $1-e^{-o_{g,t}}$ in that interval. Memory and fatigue update by convex combinations and therefore remain in $[0,1]$. Expected arrival innovations are zero, so common states and centered unit-specific states decay according to their retention factor.

Define $M_B=\max_l\sum_h|B_{hl}|+\|b_f\|_\infty$. Because the infinity norm of the difference between behavioral and reference rates is at most one, the response shift satisfies
\begin{equation}
\|\delta_t\|_\infty\leq
\rho_\delta^t\|\delta_0\|_\infty+
\frac{M_B(1-\rho_\delta^t)}{1-\rho_\delta}.
\end{equation}
The bound separates the decay of the initial shift from the accumulation of daily feedback along the deterministic expected-feedback path. The recurrence-error constants $L_t$ depend jointly on retention factors, feedback coefficients, and the sensitivity of the observation readout.

\section{Experimental Protocols}
\label{app:protocol}

\subsection{Real-Data Preprocessing}

\input{tables/data-windows.tex}

Retail preprocessing removes duplicate rows, cancellations, nonpositive quantities or prices, non-merchandise entries, and conflicting records within invoices, retaining 39,434 positive-price merchandise invoices. Each annual window uses the preceding December for warm-up, January--February for training, March 1--14 for validation, and the remaining March dates for testing.

KuaiRand experiments retain recommendation-page logs (\texttt{tab=1}) from the candidate video pool. The standard-recommendation log records videos selected by the platform's recommendation policies. The random-exposure log records videos that KuaiRand inserted into recommendation lists with a fixed probability, each replacing one recommended video with a video sampled uniformly from the candidate pool \citep{gao2022kuairand}. To assign users to buckets, we interpret the first eight bytes of the SHA-256 hash of each user identifier as a big-endian integer and take its remainder modulo 64. The main aggregate experiment uses bucket 1. The six previously fitted event-supervised models used for the analytic transformation in Section~\ref{sec:numerical-results} come from bucket 0. The two samples contain disjoint users and use the corresponding calendar windows.

Event dates are recalculated from timestamps in the Asia/Shanghai time zone. Standard recommendation uses warm-up through April 12, 2022; training on April 13--21; validation on April 22--28; and testing on April 29--May 8. Random exposure uses warm-up through April 24; training on April 25--28; validation on April 29--May 1; and testing on May 2--8.

Cohort assignments are fixed before training. For standard recommendation, history through April 12 determines the cohorts and cohort reference sizes. For random exposure, these are determined from standard-recommendation history on April 8--21. Historical click rates are smoothed with 20 reference exposures and divided into four quantile groups, then crossed with two quantile groups of historical exposure volume to produce eight cohorts. New users enter a ninth cohort. The total reference sizes of the nine cohorts are 354 and 381 for standard recommendation and random exposure, respectively, including one smoothing unit per cohort. Observation units comprise the four creators with the most warm-up exposures and a pooled unit for all remaining creators.

The aggregate model's history summaries are updated only from previously released pattern counts; the event-level model additionally uses individual histories preceding exposures. The exponential moving averages of aggregate counts use retention 0.5 for KuaiRand and 0.85 for Retail. Compared models within each task share these summaries.

Retail 2010 and 2011 contain 1,665 and 1,549 warm-up invoices, respectively. The three known cohorts separate customers by purchase frequency and spending, as in RFM segmentation \citep{fader2005rfm,chen2012data}. Reference customer counts for these cohorts are 671, 140, and 140 in 2010 and 622, 131, and 131 in 2011. The cohort for customers absent from warm-up or missing customer identifiers uses one fixed reference unit. Warm-up medians of merchandise value, quantity, and distinct-product count are \pounds294.52, 135, and 14 in 2010 and \pounds254.40, 110, and 13 in 2011. Daily labels indicate values strictly above the corresponding thresholds. Dates with no transactions remain in the sequence with zero counts.

\subsection{Model Evaluation Setup}

The aggregate experiments use seeds 20260915, 20260916, and 20260917. The controlled choice and dynamics experiments use seeds 20260912, 20260913, and 20260914. DeepAR-joint is evaluated in all four real-data windows, and the GRU predictors in the two KuaiRand windows. The full-projection GRU has 565 trainable parameters and is fitted under a separate supplementary protocol.

The main text reports each model's test results after validation-based epoch selection. The full-projection GRU uses a fully trainable hidden-state projection for every behavior. Its supplementary comparison was conducted after some test results for the original candidates were available, and its epochs were selected on validation data.

Behavioral-pattern NLL is conditional on a recorded exposure or retained invoice. Multistep behavioral evaluation uses cohort context frozen at the origin. Behavioral-count models and baselines are compared by MAE at identical forecast origins and endpoints. Longer horizons have fewer eligible origins within a fixed window; Tables~\ref{tab:retail-counts} and~\ref{tab:dynamic-ablation} and Appendix~\ref{app:behavior-scoring} give the corresponding counts.

\paragraph{DeepAR implementation and selection.}
DeepAR-joint retains the shared LSTM \citep{hochreiter1997long}, count scaling, negative-binomial likelihood, and sampled autoregressive prediction of \citet{salinas2020deepar}, and adds a joint-mark head. Our local implementation uses each short training sequence in full and weights cells uniformly in the count objective. Each cell is an observation-unit--cohort pair in KuaiRand or a cohort in Retail. Inputs concatenate the existing covariates, log-transformed cohort reference size and activity effort, a one-hot cell identifier, and the preceding count divided by $s_c=1+\operatorname{mean}_{\mathrm{train}}(N_c)$, where $N_c$ is cell $c$'s daily arrival count. Continuous input centers and scales are estimated on training data only. The initial previous count is zero, and subsequent training and validation steps use observed previous counts. A shared one-layer LSTM outputs NB2 mean $m_{\mathrm{NB}}=s_c\operatorname{softplus}(a)$ and dispersion $\alpha_{\mathrm{NB}}=\operatorname{softplus}(b)/\sqrt{s_c}$, adding $10^{-8}$ to each positive output before rescaling. The conditional variance is $m_{\mathrm{NB}}+\alpha_{\mathrm{NB}}m_{\mathrm{NB}}^2$.

The joint-mark head models $p(y_h\mid y_{<h},\mathrm{state})$ with state-dependent logits and learned lower-triangular mark dependencies. Enumerating binary patterns gives a normalized joint distribution. Initial mark biases use training marginal rates. The training objective averages daily event-normalized mark NLL and adds 0.05 times mean cell count NLL and $10^{-3}$ times mean squared trainable parameters. Adam uses learning rate 0.01 and gradient-norm clipping at 10. Validation selects the epoch and hidden width from 8 and 16 under the unregularized objective. The 24 DeepAR-joint candidate fits (four windows, three seeds, and two widths) yield 12 selected models before test scoring. The supplementary protocol was specified after the original Gaussian and GRU test results were available; test predictions were produced after model selection, with no subsequent tuning.

\paragraph{DeepAR forecast protocol.}
At each test origin, the recurrent state contains only preceding observations. Non-calendar covariates, cohort reference size, and activity effort are frozen at their origin values; calendar coordinates advance with the date. Multistep predictions integrate 2,048 sampled negative-binomial count histories and average their conditional endpoint means. Conditional joint-mark probabilities average sampled-history probabilities weighted by their endpoint arrival means. After scoring an origin, the observed count enters the recurrent history for the next origin. One-day NLL weights each date's loss by its observed arrivals; behavioral-count MAE averages absolute daily errors over behaviors. Table means average three separately computed seed metrics. Supplementary results retain per-seed scores, daily predictions, input and checkpoint hashes, width and epoch selections, and forecast settings.

\subsection{Numerical Comparisons}

Representation comparisons match the number of trainable distribution parameters; numerical comparisons hold fitted parameters fixed and vary quadrature precision. The real-data Gaussian and matched discrete representations use two Gaussian components and three support points, respectively, with five effective distribution parameters per cohort. The KuaiRand numerical comparison includes four aggregate Gaussian variants (DGD, fixed Gaussian, independently trained separated readout, and no internal dynamics) and DGD trained on individual events. Retail includes DGD, fixed Gaussian, and no internal dynamics. Each configuration covers two windows and three training seeds. One-dimensional Gaussian training uses seven quadrature nodes. Numerical reevaluation holds fitted parameters fixed and uses 25 nodes; the analytic-equivalence evaluation additionally uses 61 nodes.

The independently fitted Gaussian representation in the controlled choice experiment has two dimensions and two components, with 11 effective distribution parameters per cohort. Gaussian integration uses seven nodes per dimension. Choice models use Adam at learning rate 0.03, a parameter-normalized quadratic penalty weighted by $10^{-4}$, a training budget of 180 epochs, and early-stopping patience 30. Only distribution and choice-readout parameters enter this optimization. The controlled dynamics experiment uses a one-dimensional, two-component Gaussian mixture, nine quadrature nodes, dispersion 80, a training budget of 150 epochs, and early-stopping patience 25.

\section{Behavioral Scoring and Forecast Settings}
\label{app:behavior-scoring}

This appendix defines the historical baselines and behavioral scores used in the main text. Neural models use the three training seeds described there, and historical baselines are deterministic. Comparisons share data windows and forecast origins.

\subsection{Historical Joint Distributions}

Write $c=(j,g)$ for an observation-unit--cohort cell, and let $C_{ct}(y)$ be its count of behavioral pattern $y$ on date $t$, with $H$ behavioral dimensions. Training counts define a smoothed prior:
\begin{equation}
\bar q_y=\frac{\sum_{c,t\in\mathrm{train}}C_{ct}(y)+0.5}
{\sum_{c,t\in\mathrm{train},y'}C_{ct}(y')+0.5\,2^H},
\qquad
q_{ct}(y)=\frac{S_{ct}(y)+\tau\bar q_y}{\sum_{y'}S_{ct}(y')+\tau}.
\end{equation}
Training-period history uses the sum of training pattern counts as $S$ and predicts arrivals with the mean daily training count. Last-day history uses the most recently observed day's pattern counts and total count. Smoothed history exponentially averages daily pattern counts \citep{hyndman2008forecasting} and predicts arrivals with their sum. Its retention factor is $2^{-1/t_{1/2}}$, where $t_{1/2}$ is the half-life; zero-arrival dates also enter the update. Multistep predictions hold the distribution and arrival forecast fixed at the origin.

Each method selects smoothing strength from $\tau\in\{1,10,100\}$ on validation data. Smoothed history also selects its half-life from $t_{1/2}\in\{1,3,7\}$ days. Selection minimizes event-weighted validation joint NLL. Table~\ref{tab:history-selection} gives the selected retail settings; test scores are computed after selection. Each test date is predicted before its observations enter the history for the next origin. Neural models retain their selected checkpoints, whose original validation objective combines behavioral loss with weighted count loss. The supplementary behavioral evaluation was conducted after the original test windows had been analyzed; the added historical baselines follow the validation-selection rule above.

\renewcommand{\thetable}{E\arabic{table}}
\renewcommand{\theHtable}{E.\arabic{table}}
\setcounter{table}{0}
\begin{table}[htbp]
  \centering
  \caption{Validation-selected settings for historical joint distributions in Retail.}
  \label{tab:history-selection}
  \small
  \setlength{\tabcolsep}{7pt}
  \renewcommand{\arraystretch}{1.12}
  \begin{tabular}{@{}lrrrr@{}}
    \toprule
    Window & Training $\tau$ & Last-day $\tau$ & Smoothed $\tau$ & Half-life (days) \\
    \midrule
    Retail 2010 & 10 & 100 & 1 & 7 \\
    Retail 2011 & 100 & 100 & 1 & 7 \\
    \bottomrule
  \end{tabular}
\end{table}

\subsection{Probabilities, Behavioral Rates, and Counts}

Let $N_{ct}=\sum_y C_{ct}(y)$, $Y_{cth}=\sum_y C_{ct}(y)y_h$, and $p_{cth}=\sum_y q_{ct}(y)y_h$. Event-weighted joint NLL and the Brier score \citep{brier1950verification} for behavior $h$ are
\begin{equation}
\begin{aligned}
\mathrm{NLL}&=-\frac{\sum_{c,t,y}C_{ct}(y)\log q_{ct}(y)}{\sum_{c,t}N_{ct}},\\
\mathrm{Brier}_h&=\frac{\sum_{c,t}\left[Y_{cth}(1-p_{cth})^2+(N_{ct}-Y_{cth})p_{cth}^2\right]}{\sum_{c,t}N_{ct}}.
\end{aligned}
\end{equation}
Both are strictly proper scoring rules \citep{gneiting2007strictly}. Mean Brier weights the $H$ behaviors equally. Summing over cells gives observed behavioral counts, full predicted counts, and conditional predictions evaluated at observed arrival volumes:
\begin{equation}
Y_{th}=\sum_c Y_{cth},\qquad
\widehat Y_{th}=\sum_c\widehat N_{ct}p_{cth},\qquad
\widehat Y^{\mathrm{cond}}_{th}=\sum_c N_{ct}p_{cth}.
\end{equation}
Behavioral-count MAE is $H^{-1}\sum_h\operatorname{mean}_t|\widehat Y_{th}-Y_{th}|$. Conditional behavioral rates use $\widehat Y^{\mathrm{cond}}$ to evaluate response probabilities at the observed arrival composition. Observed arrival volumes enter conditional scoring; full count predictions use $\widehat N$.

Overall behavioral-rate MAE divides $100|\widehat Y^{\mathrm{cond}}_{th}-Y_{th}|$ by the total arrivals on that date, then averages over behaviors and dates with arrivals. Cohort-level scoring first pools observation units within each cohort, computes its mean date--behavior error in the same way, and then weights cohorts equally. Zero-arrival dates remain in count errors and are excluded from rate scores only when the corresponding rate denominator is zero.

\subsection{Co-occurrence Rates and Forecast Horizons}

For each behavioral pair $(h,h')$, the quantity $\sum_y q_{ct}(y)y_h y_{h'}$ predicts the probability that two behaviors occur within the same arrival. We weight these probabilities by observed arrival counts, compute daily co-occurrence-rate MAE, and average equally across all behavioral pairs. KuaiRand has 21 pairs and Retail has three.

Let $\Delta$ denote the forecast horizon. Multistep scoring evaluates the endpoint at day $\Delta$, pairing each origin's prediction with the observation at the same endpoint. One-day scores use 10, 7, 17, and 17 origins in standard recommendation, random exposure, Retail 2010, and Retail 2011, respectively. The three-day random-exposure ablation in the main text uses five origins. Neural scores are computed separately for each training seed before their means are reported; seed standard deviations describe training variation within the same window.

\section{Numerical Evaluations of Aggregate Learning}
\label{sec:numerical-results}

Following the properties derived in Appendices~\ref{app:equivalence}--\ref{app:numerical}, numerical evaluations examine likelihood computation from counts under shared conditioning information, parameter transformations, and quadrature accuracy. We also check the numerical accuracy of gradients used in joint training and their propagation across days.

Under the shared conditioning information in Appendix~\ref{app:equivalence}, the numerical difference between event and count objectives is zero after omitting the parameter-independent combinatorial constant, with a maximum gradient difference of $5.55\times10^{-17}$. In the gradient checks, central differences agree with automatic differentiation for mixture weights, means, and Cholesky parameters. Outputs across days have nonzero gradients with respect to retention, observation-correction, and feedback parameters.

The Gaussian tilting transformation in Appendix~\ref{app:equivalence} maps an arrival-weighted population distribution to a corresponding separated readout. We apply this transformation to six sets of Gaussian model parameters trained on individual events, hold the fitted parameters fixed, and compare predictions before and after transformation. With 7, 25, and 61 quadrature nodes, the maximum absolute differences in mean test NLL are $9.65\times10^{-4}$, $1.01\times10^{-5}$, and $4.36\times10^{-8}$, respectively. As integration accuracy increases, the predictions of the two parameterizations converge, consistent with the analytic equivalence under exponential tilting of a Gaussian mixture.

Equation~\eqref{eq:quadrature-bound} describes the effect of integration error on predicted probabilities. We increase quadrature from the seven nodes used in training to 25 nodes, hold fitted parameters fixed, and recompute test behavioral losses. The KuaiRand comparison covers 30 fitted Gaussian models: four aggregate Gaussian variants and one individual-event training configuration, each with two windows and three seeds. The maximum absolute change in test NLL is 0.001747. For the standard-recommendation DGD models in Table~\ref{tab:representation}, the maximum change is 0.000251. Retail includes three Gaussian variants across two windows and three seeds; the maximum change across these 18 models is $1.62\times10^{-5}$. Appendix~\ref{app:protocol} lists the variants and numerical settings.

\input{sections/appendix-training}

\section{Additional Related Work}
\label{app:related}

Hawkes processes model self-exciting event intensities and predict the popularity of social-media posts \citep{hawkes1971spectra,zhao2015seismic}. Probabilistic forecasters learn count distributions over time with recurrent or deep state-space models \citep{salinas2020deepar,rangapuram2018deep}, and count autoregressions let past counts drive the conditional mean of later counts \citep{ferland2006ingarch,fokianos2009poisson}. Recommender simulators generate user responses for evaluating recommendation policies \citep{ie2019recsim,zhao2023kuaisim}, and SARN disaggregates coarse spatio-temporal counts to finer levels \citep{han2024sarn}. DGD predicts such count sequences through a shared response distribution, contact-weighted aggregation, and feedback recurrence.

%% file: tables/data-windows.tex
\begin{table}[!htbp]
  \centering
  \caption{Real-data windows. Entries separated by slashes give training, validation, and test statistics. Events are exposures in KuaiRand and retained invoices in Retail.}
  \label{tab:data}
  \small
  \setlength{\tabcolsep}{4pt}
  \renewcommand{\arraystretch}{1.12}
  \begin{tabular}{@{}lrrr@{}}
    \toprule
    Window & Days & Events & Behaviors \\
    \midrule
    Standard & 9 / 7 / 10 & 4,655 / 1,251 / 1,881 & 7 \\
    Random & 4 / 3 / 7 & 2,256 / 2,033 / 9,101 & 7 \\
    Retail 2010 & 59 / 14 / 17 & 2,237 / 682 / 960 & 3 \\
    Retail 2011 & 59 / 14 / 17 & 2,159 / 565 / 869 & 3 \\
    \bottomrule
  \end{tabular}
\end{table}

%% file: sections/appendix-training.tex
\clearpage
\section{Joint Training and Detailed Computation}
\label{app:training}

Algorithm~\ref{alg:training} separates within-epoch state progression from parameter optimization. The observation operator $\mathcal O_\theta$ comprises Equations~\eqref{eq:population}--\eqref{eq:quadrature}; the transition operator $\mathcal T_{\theta_T}$ consumes feedback counts and predicted arrivals to calculate state updates and arrival innovations. Figure~\ref{fig:method} expands the node-level computation and the two feedback branches.

\begin{algorithm}[htbp]
\caption{Joint learning of DGD from aggregate observations}
\label{alg:training}
\begin{algorithmic}[1]
\REQUIRE Training counts $C_{jgt}(y)$, known inputs, reference sizes, validation sequence.
\STATE Initialize $\theta_P,\theta_O,\theta_T$; compute feature scales from training data.
\FOR{each optimization epoch}
  \STATE Reset the daily state $s_1$ and accumulated loss $L\gets0$.
  \FOR{each training date $t=1,\ldots,T$}
    \STATE Predict $(\widehat N_t,q_t,\widehat Y_t)\gets\mathcal O_\theta(s_t,x_t,n_t,e_t)$.
    \STATE Accumulate $L\gets L+\mathcal L_{B,t}+\alpha\mathcal L_{N,t}$.
    \IF{$t<T$}
      \STATE Form $s_{t+1\mid t}^{\mathrm{free}}\gets\mathcal T_{\theta_T}(s_t,\widehat N_t,\widehat Y_t;\widehat N_t)$.
      \STATE Fix origin history summaries and sizes; advance known calendar and effort inputs.
      \STATE Predict from $s_{t+1\mid t}^{\mathrm{free}}$ and score the next day's aggregate labels.
      \STATE Accumulate $L\gets L+\beta\mathcal L_{t+1\mid t}^{\mathrm{free}}$.
    \ENDIF
    \STATE Advance $s_{t+1}\gets\mathcal T_{\theta_T}(s_t,N_t,Y_t;\widehat N_t)$ using observed feedback.
  \ENDFOR
  \STATE Form $\mathcal L\gets L/T+\lambda_R\mathcal R(\theta)$.
  \STATE Backpropagate through quadrature and the complete sequence; clip gradients.
  \STATE Update $\theta_P,\theta_O,\theta_T$ jointly using Adam.
  \STATE Select saved parameters and stopping epoch using the validation objective.
\ENDFOR
\end{algorithmic}
\end{algorithm}

\clearpage
\section{Daily Retail Behavioral-Count Forecasts}
\label{app:trajectories}

\begin{figure}[htbp]
\centering
\includegraphics[width=\linewidth]{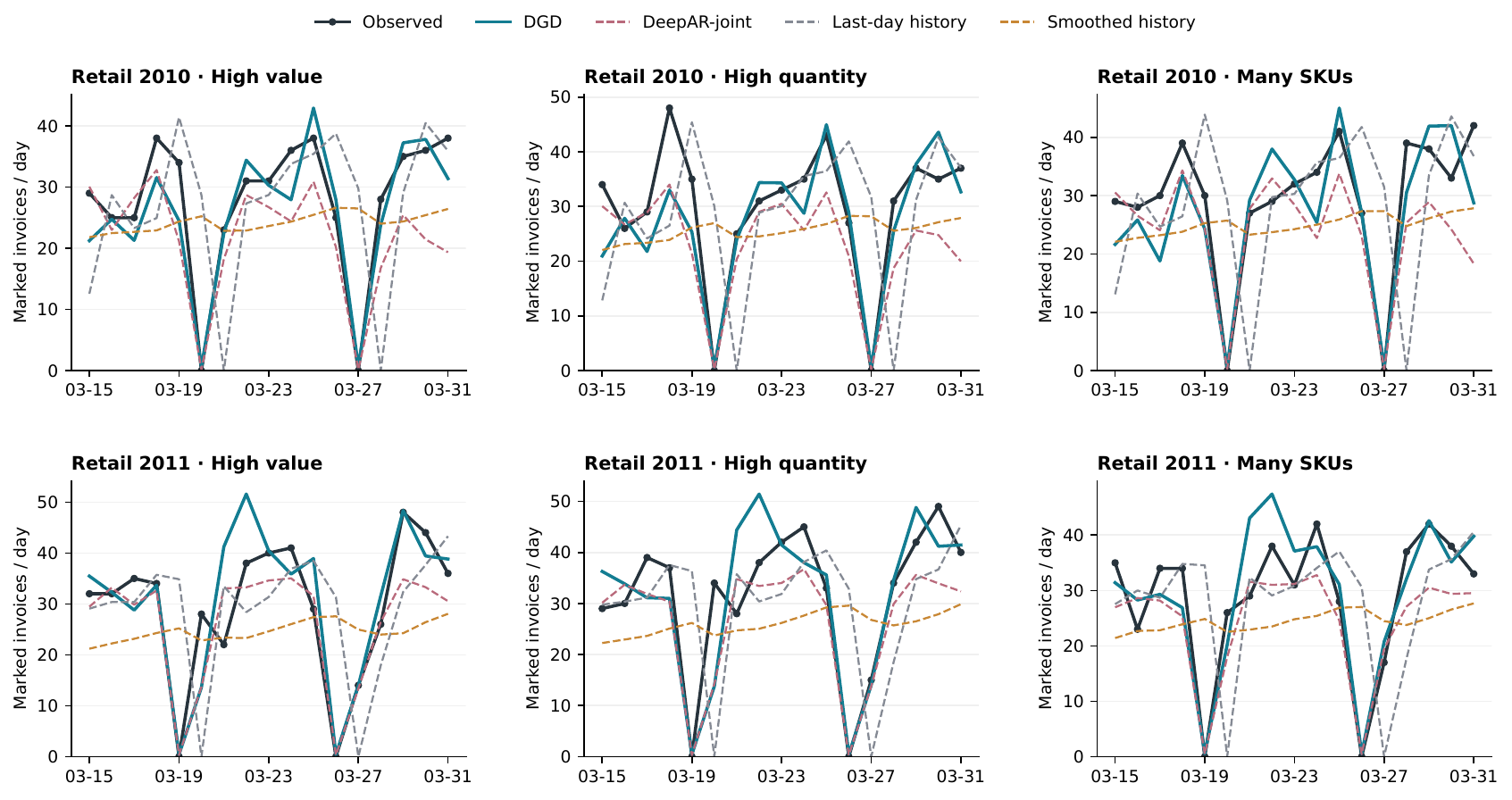}
\caption{Daily behavioral-count forecasts in both retail windows. Columns show invoice counts for high value, high quantity, and many SKUs. Each forecast uses preceding observations and combines predicted arrivals with behavioral probabilities. Curves average predictions across three training seeds. Table~\ref{tab:retail-counts} averages the errors computed separately for each seed.}
\label{fig:retail-behavior}
\end{figure}